%% file: arxiv.tex
\documentclass[letterpaper]{article} % DO NOT CHANGE THIS
\usepackage[preprint]{aaai2027}  % DO NOT CHANGE THIS
\usepackage[hyphens]{url}  % DO NOT CHANGE THIS
\usepackage{graphicx} % DO NOT CHANGE THIS
\usepackage{natbib}  % DO NOT CHANGE THIS AND DO NOT ADD ANY OPTIONS TO IT
\usepackage{caption} % DO NOT CHANGE THIS AND DO NOT ADD ANY OPTIONS TO IT
\usepackage{algorithm}
\usepackage{algorithmic}
\usepackage{amsmath}
\usepackage{booktabs}
\usepackage{multirow}
\usepackage{graphicx}

\usepackage{newfloat}
\usepackage{listings}
\DeclareCaptionStyle{ruled}{labelfont=normalfont,labelsep=colon,strut=off} % DO NOT CHANGE THIS
\floatstyle{ruled}
\newfloat{listing}{tb}{lst}{}
\floatname{listing}{Listing}

\usepackage{booktabs}

\title{RACO: Reliability-Aware Coarse-Goal Optimization for Inspection-Oriented UAV Vision-Language Navigation}
\author{
    Sen Wang\textsuperscript{\rm 1},
    Yiming Sun\textsuperscript{\rm 1},
    Jiaxuan He\textsuperscript{\rm 2},
    Pengfei Zhu\textsuperscript{\rm 1}\corresponding
}

\affiliations{
    \textsuperscript{\rm 1}School of Automation, Southeast University, Nanjing, China\\
    \textsuperscript{\rm 2}School of Science and the School of Engineering,\\
    The Hong Kong University of Science and Technology, Hong Kong\\
    Corresponding author: zhupengfei@tju.edu.cn
}

\begin{document}

\maketitle

\begin{abstract}
UAV vision-language navigation (UAV-VLN) is commonly evaluated as goal reaching, but inspection-oriented deployment requires the agent to stop within a valid inspection region and avoid falsely confirming visually or semantically similar distractors. This requirement exposes a key weakness in existing coarse-to-fine UAV-VLN policies: the coarse goal predicted before local refinement is often treated as reliable, although it may drift toward plausible but incorrect object regions and limit the ability of the local stage to recover. To systematically evaluate this problem, we introduce LG-UVI, an object-centric inspection evaluation setting derived from CityNav/CityRefer. LG-UVI extends standard UAV-VLN episodes with target objects, hard distractors, type-aware inspection regions, and diagnostics for inspection-region arrival and object-level confirmation. To address this inspection-oriented setting, we further propose RACO, a reliability-aware adaptive coarse-to-fine navigation framework. Instead of treating the predicted coarse goal as a fixed waypoint, RACO views it as a runtime hypothesis and uses object-level candidate anchors to check and correct coarse localization before Stage~1 and at the Stage~1-to-Stage~2 boundary. RACO also applies scale-adaptive terminal refinement to handle terminal near-miss cases using runtime-observable geometric and anchor-based evidence. Under a unified online evaluation protocol, RACO improves SR over the reproduced HETT baseline by 9.53 and 7.98 percentage points on validation-unseen and test-unseen, respectively. It also improves inspection-region arrival and reduces false verification risk, showing that coarse-goal reliability optimization is an effective complement to existing coarse-to-fine UAV-VLN policies.
\end{abstract}


\begin{figure*}[t]
  \centering
  \includegraphics[width=0.8\textwidth]{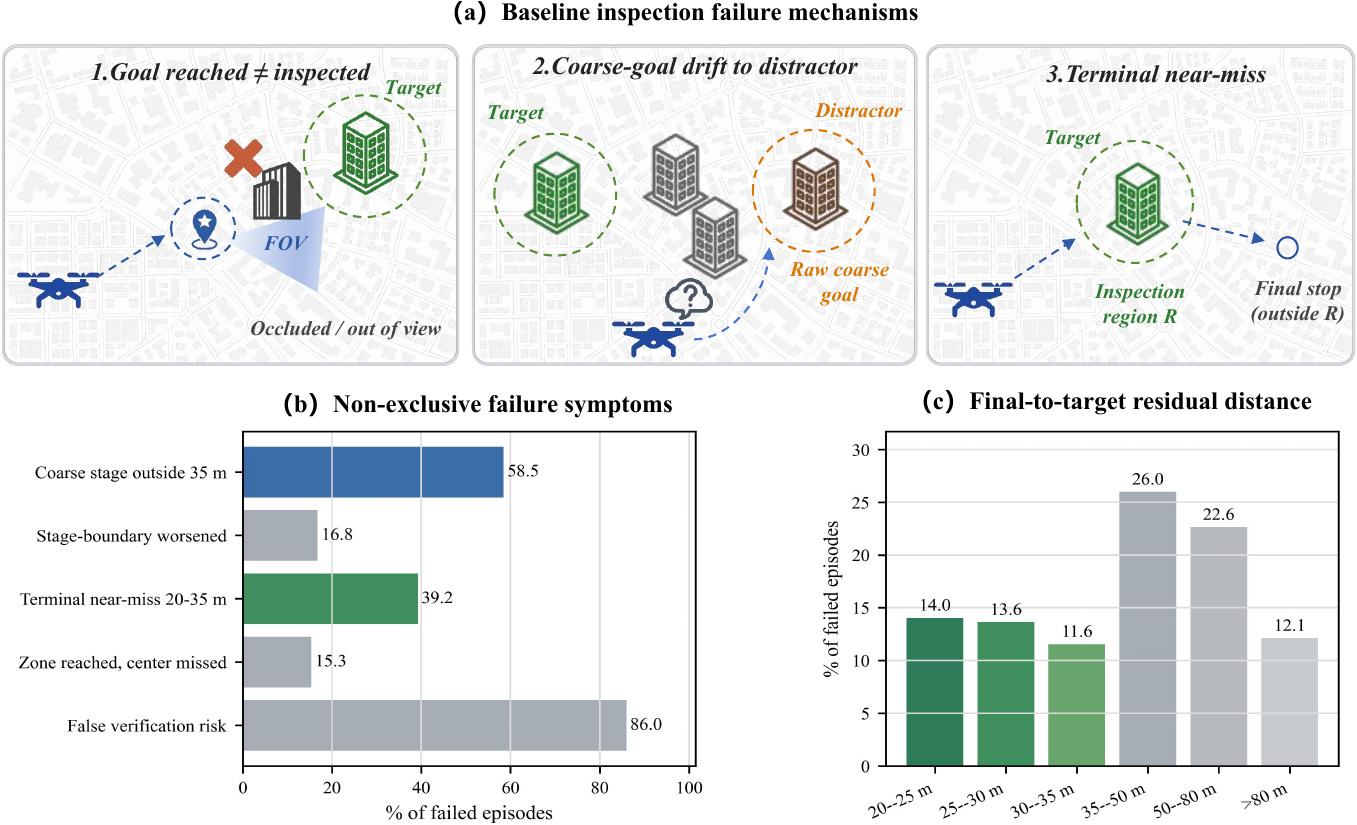}
  \caption{
Failure diagnosis for inspection-oriented UAV-VLN.
(a) Typical HETT failures: invalid inspection states, coarse-goal drift to same-category distractors, and terminal near-misses.
(b) Non-exclusive failure symptoms among failed episodes.
(c) Final-to-target residual distances, with the 20--35 m near-miss band highlighted.
Ground-truth distances are used only for offline diagnosis.
}
  \label{fig:motivation_failure_diagnosis}
\end{figure*}

\section{Introduction}

Vision-language navigation (VLN) requires an embodied agent to follow natural-language instructions by grounding them in visual observations and actions~\cite{anderson2018vision}. Recent work has extended VLN from indoor environments to city-scale aerial navigation, where UAVs interpret instructions and execute long-range trajectories~\cite{liu2023aerialvln,lee2025citynav,miyanishi2023cityrefer}. Existing UAV-VLN benchmarks primarily evaluate coordinate-level goal reaching using navigation error (NE), success rate (SR), and success weighted by path length (SPL)~\cite{liu2023aerialvln,lee2025citynav,wang2025traveluav,ding2026hett,gao2026openfly}. However, these metrics do not fully capture the requirements of inspection-oriented deployment.

In an inspection task, reaching the vicinity of a target coordinate is not enough. The agent must stop in a suitable viewing region, maintain a final state that supports inspection of the intended object, and avoid confirming nearby objects with similar visual or semantic cues. This distinction is important in dense urban scenes, where buildings, cars, parking areas, and ground regions often appear close to one another and may belong to the same category. A UAV may reach the correct neighborhood while still failing to produce a valid inspection state, or it may associate its final state with a same-category distractor. Inspection-oriented UAV-VLN therefore exposes reliability failures that standard goal-reaching metrics tend to hide.

Fig.~\ref{fig:motivation_failure_diagnosis} illustrates this gap through a diagnosis of the reproduced HETT baseline~\cite{ding2026hett} under the inspection-oriented evaluation protocol used in this work. HETT failures often stem from unreliable coarse localization, drift toward plausible same-category distractors, or terminal near-misses where the agent stops just outside the valid inspection region. These failures are not isolated. As shown in Fig.~\ref{fig:motivation_failure_diagnosis}, many failed episodes involve coarse-goal reliability issues, and a substantial fraction falls into the 20--35 m terminal near-miss band. The distance statistics in Fig.~\ref{fig:motivation_failure_diagnosis} are used only to diagnose the reproduced baseline offline.

To systematically study these challenges, we introduce LG-UVI, an object-centric inspection evaluation setting derived from CityNav and CityRefer~\cite{lee2025citynav,miyanishi2023cityrefer}. LG-UVI preserves the original UAV-VLN episodes while augmenting them with target objects, same-category candidates, hard distractors, and type-aware inspection regions. Beyond standard metrics such as NE, SR, and SPL, LG-UVI evaluates whether the agent reaches a valid inspection region, confirms the intended object, and avoids false verification of nearby hard distractors. This setting separates coordinate-level navigation success from inspection-region arrival, object-level confirmation, and false verification risk.

We further propose RACO, a reliability-aware coarse-goal optimization framework for inspection-oriented UAV-VLN. RACO keeps the original two-stage navigation backbone, but treats the predicted coarse goal as a runtime hypothesis rather than a fixed waypoint. It verifies and corrects this hypothesis using object-level candidate anchors before Stage~1 and at the transition from Stage~1-to-Stage~2, reducing the chance that an unreliable coarse goal is passed into local refinement. RACO also introduces scale-adaptive terminal inspection refinement to repair bounded near-miss errors using only runtime-observable geometric and anchor-based features. 

Experiments on LG-UVI show that RACO consistently improves the reproduced HETT baseline under the same online evaluation protocol. On the validation-unseen and test-unseen splits, RACO increases SR by 9.53 and 7.98 percentage points, respectively, while also improving path efficiency. It further improves inspection-region arrival and reduces false verification risk, indicating that reliability-aware coarse-to-fine correction is useful beyond coordinate-level navigation success.

Our contributions are summarized as follows:
\begin{itemize}
  \item We introduce LG-UVI, an inspection-oriented extension of CityNav/CityRefer for object-level UAV inspection. LG-UVI preserves standard UAV-VLN episodes while augmenting them with target objects, same-category candidates, hard distractors, type-aware inspection regions, and diagnostics for inspection-region arrival, object confirmation, and false verification.

  \item We propose a stage-aware coarse-goal reliability correction mechanism. Instead of directly trusting the raw coarse goal predicted by a two-stage UAV-VLN policy, our method treats it as a runtime hypothesis and verifies it with object-level candidate anchors before Stage~1 and at the Stage~1-to-Stage~2 boundary, reducing drift toward plausible but incorrect object regions.

  \item We develop a scale-adaptive terminal inspection refinement module and integrate it with coarse-goal correction into RACO. The resulting framework addresses bounded terminal near-miss cases using only runtime-observable features and consistently improves the HETT baseline under a unified online evaluation protocol.
\end{itemize}

\begin{table*}[t]
\centering
\small
\begin{tabular}{lrrrrrrrl}
\toprule
Split & Episodes & Building & Car & Ground & Parking & Avg. Cand. & Avg. Hard Distr. & Zone Policy \\
\midrule
Train Seen   & 21,878 & 8,268 & 11,641 & 1,226 & 743 & 30.84 & 14.73 & center / contour \\
Val Seen     & 2,470  & 1,057 & 1,226  & 131   & 56  & 30.88 & 14.84 & center / contour \\
Val Unseen   & 2,697  & 1,141 & 1,026  & 392   & 138 & 29.71 & 14.58 & center / contour \\
Test Unseen  & 5,281  & 2,750 & 1,693  & 631   & 207 & 30.55 & 14.74 & center / contour \\
\bottomrule
\end{tabular}
\caption{
Statistics of LG-UVI. Each split preserves the original CityNav/CityRefer navigation episodes and adds object-centric inspection annotations. Candidate pools contain the target object and same-category alternatives, while hard distractors are the nearest same-category objects used for false-verification diagnosis. Cars use center-radius inspection regions, whereas buildings, ground regions, and parking areas use contour-buffer inspection regions.
}
\label{tab:lguvi_statistics}
\end{table*}

\section{Related Work}

\subsection{UAV Vision-Language Navigation}

Vision-language navigation (VLN) asks an embodied agent to follow natural-language instructions by grounding them in visual observations and action decisions. Early indoor benchmarks and methods, including R2R~\cite{anderson2018vision}, HAMT~\cite{chen2021history}, DUET~\cite{chen2022think}, data-scaling approaches~\cite{wang2023scaling}, and volumetric representations~\cite{liu2024volumetric}, mainly evaluate whether an agent can reach a navigation goal efficiently. These studies established standard metrics such as navigation error, success rate, and path efficiency, but they were largely designed for ground-level indoor environments.

Recent work has extended VLN to aerial agents and city-scale outdoor scenes. AerialVLN~\cite{liu2023aerialvln} introduces language-guided UAV navigation in outdoor 3D environments. CityNav~\cite{lee2025citynav} and CityRefer~\cite{miyanishi2023cityrefer} provide city-scale language, trajectory, and object grounding resources for aerial navigation. TravelUAV~\cite{wang2025traveluav} and OpenFly~\cite{gao2026openfly} further improve the realism and scale of UAV-VLN evaluation through more complete simulation platforms and benchmark settings. Recent methods also explore stronger reasoning, memory, and foundation-model-based planning for aerial navigation, including HETT~\cite{ding2026hett}, CityNavAgent~\cite{zhang2025citynavagent}, FlightGPT~\cite{cai2025flightgpt}, See, Point, Fly~\cite{hu2025see}, OpenVLN~\cite{lin2025openvln}, LongFly~\cite{jiang2025longfly}, and AutoFly~\cite{sun2026autofly}. Most existing approaches nevertheless retain coordinate-level success criteria and do not explicitly evaluate inspection-region arrival or confusion with same-category distractors.

\subsection{Object-Centric Inspection and Reliability}

Object-centric embodied tasks require agents to reason about semantic categories, object instances, and task-specific grounding rather than only metric goal locations. REVERIE~\cite{qi2020reverie} studies remote object grounding during embodied navigation, while ALFRED~\cite{shridhar2020alfred} requires grounded instruction following and object interaction. Language-conditioned robotic systems and vision-language-action models further connect semantic instructions with action affordances and embodied control~\cite{ahn2023can,shah2023lm,driess2023palme,zitkovich2023rt}. A related line of work improves robustness through self-monitoring, backtracking, and replanning: progress monitors estimate navigation progress~\cite{ma2019selfmonitoring}, regretful and tactical agents learn when to backtrack or rewind from poor local decisions~\cite{ma2019regretful,ke2019tactical}, and introspective or VLM/LLM-based planners predict failures and adapt plans under changing observations~\cite{rabiee2022competence,skreta2024replan,mei2024replanvlm}. RACO is related to these ideas, but focuses on coarse-goal reliability in two-stage UAV-VLN, grounding correction in object-level inspection anchors rather than generic progress estimation or backtracking over visited viewpoints.

Aerial domains are also moving toward application-oriented language-conditioned execution, including open-vocabulary grounding, terminal delivery, mission generation, onboard aerial navigation, and interactive aerial navigation~\cite{zhang2025vlfly,zhang2025logisticsvln,sautenkov2025uav,wu2025vla,chen2026aerialvla}. Our work addresses a complementary problem: inspection-oriented reliability in two-stage UAV-VLN. In dense urban scenes, a UAV can reach the correct neighborhood while still stopping outside the valid inspection region or associating its final state with a nearby distractor. LG-UVI separates inspection-region arrival from object-level confirmation, while RACO treats the coarse goal as a runtime hypothesis and revises unreliable coarse and terminal states through object-level candidate anchors.

\begin{figure*}[t]
\centering
\includegraphics[width=0.8\textwidth]{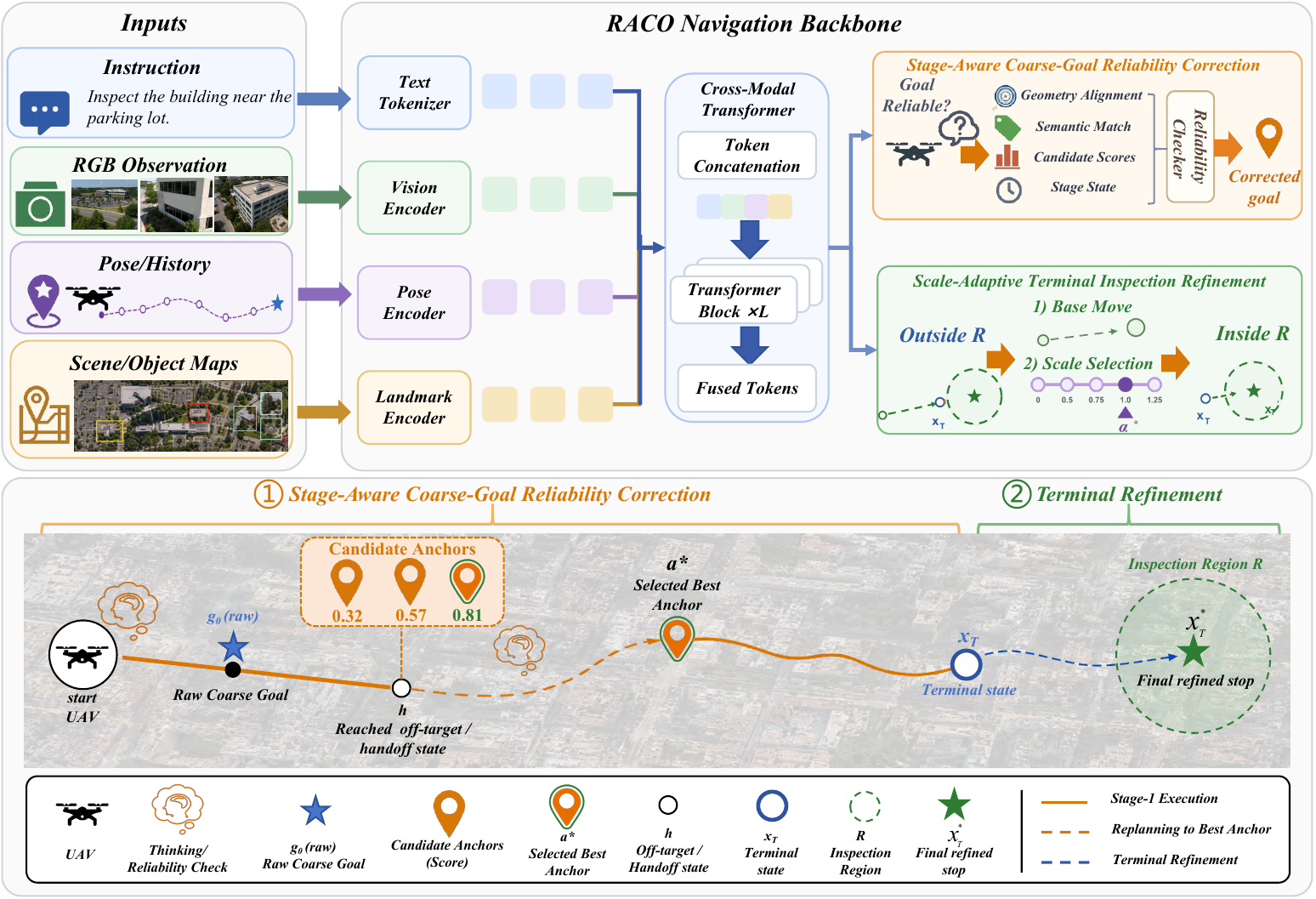}
\caption{
Overview of the proposed RACO framework. RACO augments a two-stage UAV-VLN backbone with stage-aware coarse-goal reliability correction and scale-adaptive terminal inspection refinement, and the lower panel visualizes the resulting correction process in one representative episode.
}
\label{fig：overview}
\end{figure*}

\section{LG-UVI Benchmark}

LG-UVI extends CityNav/CityRefer from coordinate-level UAV-VLN to inspection-oriented navigation. In each episode, an aerial agent follows a language instruction $I$ from an initial state $s_0$ and stops at $s_T$ to inspect a target object $o^\ast$ in a 3D urban scene. Unlike standard UAV-VLN, where success is mainly determined by distance to a target coordinate, inspection requires the agent to stop in a suitable viewing region and associate its final state with the intended object. This distinction matters in urban environments, where multiple same-category objects may appear close to one another and visually plausible distractors can lead to incorrect inspection.

LG-UVI preserves the original instructions, trajectories, maps, start poses, and target positions from CityNav/CityRefer, and adds object-centric inspection metadata to each episode. Each target object is represented by its object ID, semantic type, geometry, and a type-aware inspection region $Z(o^\ast)$. The main object types include buildings, cars, ground regions, and parking areas. Each episode also includes same-category candidate objects and hard distractors, allowing the evaluator to test whether the agent reaches the correct inspection region and avoids plausible but incorrect targets. The candidate pool contains the ground-truth target object, the nearest same-type candidates, and deterministic additional same-type negatives, with the pool capped at 32 objects. Hard distractors are defined as the nearest same-type objects according to 2D object distance. The inspection region is type-aware: cars use a 20 m center-radius region, while buildings, ground regions, and parking areas use a 20 m contour-buffer region. Table~\ref{tab:lguvi_statistics} summarizes the resulting split statistics.

We report standard navigation metrics, including NE, SR, and SPL, and add four inspection-oriented diagnostics. Zone-SR measures whether the final position enters the type-aware inspection region, OSA measures whether the inspected object $\hat{o}$ matches the target object $o^\ast$, ISR requires both region arrival and correct object confirmation, and FVR measures whether the agent falsely verifies a hard distractor. These metrics separate coordinate-level navigation, inspection-region arrival, object-level confirmation, and false-verification risk. Formal metric definitions and the post-hoc object-binding protocol are provided in Supplementary Section~S5.

\section{Method}
\subsection{Overview of RACO}

Fig.~\ref{fig：overview} summarizes RACO, a stage-aware reliability framework that extends the HETT execution process without modifying its navigation backbone. Given a language instruction, visual observations, pose history, and a fixed scene-level object map, the backbone produces three runtime states: a raw coarse goal $g_0$, a Stage~1 handoff state $x_1$, and a terminal state $x_T$. RACO acts on these states at three points through two coordinated mechanisms. Stage-aware coarse-goal correction estimates the reliability of the current state and the utility of type-compatible object anchors from geometric, semantic, candidate-score, and stage-context features. It may replace $g_0$ before Stage~1 and, at the Stage~1-to-Stage~2 boundary, update the coarse goal and perform a bounded replan before local refinement. Scale-adaptive terminal refinement then scores a discrete set of geometry-conditioned correction actions, including abstention, to repair terminal near-misses after Stage~2. The decision modules are trained separately.

\subsection{Object-Level Candidate Anchors}

RACO introduces object-level candidate anchors to bridge coordinate-level navigation and object-level inspection. Rather than treating the predicted coarse goal as an isolated 3D point, RACO grounds it in nearby semantic objects that can provide more reliable references for subsequent navigation.

We use a fixed scene-level object map $\mathcal{M}$ derived from the CityRefer annotations, where each object is represented by its semantic category, reference position, and geometry. A lightweight language-to-type parser extracts a target-category cue $\hat{c}$ from the instruction; when no category can be identified, type filtering is disabled. Given the raw coarse goal $g_0$, RACO retrieves up to $K$ nearby objects compatible with $\hat{c}$:
\begin{equation}
\label{eq:runtime-anchor-set}
\mathcal{A}_{\mathrm{run}}(g_0,\hat{c};\mathcal{M})
=
\{a_j\}_{j=1}^{K}.
\end{equation}

Each anchor $a_j$ provides a geometry-aware reference position $p(a_j)$ together with features describing its spatial compatibility with $g_0$, semantic consistency with the instruction, and distinctiveness among neighboring candidates. Stage-specific modules score these anchors and select the references used for coarse-goal correction and terminal refinement. Additional anchor-construction details, feature groups, map-aware controls, and synthetic map perturbations are provided in Supplementary Sections~S1 and S3.

The runtime anchor set $\mathcal{A}_{\mathrm{run}}$ is constructed around the model-predicted goal and contains neither episode-specific target identities nor hard-distractor labels. It is distinct from the ground-truth-inclusive candidate pool used only for evaluation.

\subsection{Stage-Aware Coarse-Goal Reliability Correction}

RACO decomposes coarse-goal correction into two decisions: whether the current state requires intervention and which object anchor should serve as the correction target. Both decisions are made before Stage~1 and at the Stage~1-to-Stage~2 boundary using separately parameterized modules.

Before Stage~1, a reliability gate estimates an intervention probability $\rho_{\mathrm{pre}}$ from geometric, semantic, candidate-distribution, and stage-context features. In parallel, an anchor-utility scorer assigns each runtime anchor $a_j \in \mathcal{A}_{\mathrm{run}}$ a score $s_j^{\mathrm{pre}}$ and selects:
\begin{equation}
\label{eq:pre-anchor-selection}
a_{\mathrm{pre}}^\star
=
\arg\max_{a_j\in\mathcal{A}_{\mathrm{run}}}
s_j^{\mathrm{pre}}.
\end{equation}

The coarse goal is then updated as
\begin{equation}
\label{eq:pre-goal-update}
g_0'=
\begin{cases}
p(a_{\mathrm{pre}}^\star), & \text{if } \Gamma_{\mathrm{pre}}=1,\\
g_0, & \text{otherwise},
\end{cases}
\end{equation}
where $\Gamma_{\mathrm{pre}}$ indicates that the intervention probability, anchor utility, score margin, and displacement constraint jointly meet their acceptance criteria. Otherwise, Stage~1 proceeds with the original coarse goal.

After Stage~1 reaches the handoff state $x_1$, a boundary reliability gate evaluates whether the state is suitable for local refinement using the reached endpoint, executed trajectory, and local anchor distribution. A boundary-specific scorer selects:
\begin{equation}
\label{eq:boundary-anchor-selection}
a_{\mathrm{B}}^\star
=
\arg\max_{a_j\in\mathcal{A}_{\mathrm{run}}}
s_j^{\mathrm{B}}.
\end{equation}

If the handoff state is unreliable, RACO updates the coarse goal to $g_0''=p(a_{\mathrm{B}}^\star)$ and performs a bounded replan before entering Stage~2. Replanning stops when the agent reaches the selected anchor within a fixed tolerance or the preset iteration limit is reached.

The two modules are trained on train-seen episodes and kept fixed during evaluation. The learned decision modules, training objectives, acceptance criteria, and replanning parameters are detailed in Supplementary Section~S1.

\subsection{Scale-Adaptive Terminal Inspection Refinement}
Scale-adaptive terminal refinement complements the coarse-goal reliability loop by correcting bounded near-misses at the end of execution. While coarse-goal correction reduces larger localization errors before local refinement, the terminal policy adjusts final states that remain close to, but outside, a valid inspection region. The correction is performed online as part of the navigation process rather than as offline post-processing.

Let $x_T$ denote the terminal position after Stage~2, and let $a_T$ be the runtime anchor selected from the candidate anchor space. This anchor is produced by RACO during inference and is not a ground-truth target. We define the terminal residual distance as
\begin{equation}
\label{eq:terminal-residual}
d_T = \|x_T - p(a_T)\|_2,
\end{equation}
where $p(a_T)$ is the reference position of the runtime anchor. RACO first maps this residual distance to a geometry-induced base move:
\begin{equation}
\label{eq:geometry-base-move}
m_{\mathrm{geo}}(d_T) =
\begin{cases}
5,  & 20 \le d_T < 25,\\
10, & 25 \le d_T < 30,\\
15, & 30 \le d_T \le 35,\\
0,  & \text{otherwise}.
\end{cases}
\end{equation}
This base move defines a bounded correction toward the runtime anchor. No refinement is applied when $d_T$ falls outside the specified near-miss range.

Rather than executing the base move directly, RACO uses a learned policy to select a scale from a small discrete action set:
\begin{equation}
\label{eq:scale-action-set}
\alpha^\star \in \{0, 0.5, 0.75, 1.0, 1.25\},
\end{equation}
where $\alpha^\star=0$ means that the agent abstains from refinement. The refined terminal position is computed as
\begin{equation}
\label{eq:terminal-update}
x_T' =
x_T +
\alpha^\star m_{\mathrm{geo}}(d_T)
\frac{p(a_T)-x_T}{\|p(a_T)-x_T\|_2}.
\end{equation}

The discrete scale set keeps the correction simple and bounded while allowing the policy to choose a conservative or slightly larger move when supported by the runtime evidence.

The scale-selection policy uses runtime-observable features, including the terminal residual distance, the geometry-based base move, the runtime anchor score and margin, local candidate ambiguity, stage-boundary replanning status, and the geometric relation between the terminal state and the anchor. The policy is trained before evaluation, applied online after Stage~2, and executed before the final metrics are computed.

\begin{table*}[t]
\centering
\begin{tabular}{lcccccccccccc}
\toprule
Model
& \multicolumn{4}{c}{Validation Seen}
& \multicolumn{4}{c}{Validation Unseen}
& \multicolumn{4}{c}{Test Unseen} \\
\cmidrule(lr){2-5}
\cmidrule(lr){6-9}
\cmidrule(lr){10-13}
& NE$\downarrow$ & SR$\uparrow$ & OSR$\uparrow$ & SPL$\uparrow$
& NE$\downarrow$ & SR$\uparrow$ & OSR$\uparrow$ & SPL$\uparrow$
& NE$\downarrow$ & SR$\uparrow$ & OSR$\uparrow$ & SPL$\uparrow$ \\
\midrule

Random
& 222.3 & 0.00  & 1.15  & 0.00
& 223.0 & 0.00  & 0.90  & 0.00
& 208.8 & 0.00  & 1.44  & 0.00 \\

Human
& 9.1   & 89.31 & 96.40 & 60.17
& 9.4   & 88.39 & 95.54 & 62.66
& 9.8   & 87.86 & 95.29 & 57.04 \\

Seq2Seq
& 257.1 & 1.81  & 7.89  & 1.58
& 317.4 & 0.79  & 8.82  & 0.61
& 245.3 & 1.50  & 8.34  & 1.30 \\

CMA
& 240.8 & 0.95  & 9.42  & 0.92
& 268.8 & 0.65  & 7.86  & 0.63
& 252.6 & 0.82  & 9.70  & 0.79 \\

AerialVLN
& 65.6  & 9.77  & 23.77 & 8.64
& 81.8  & 6.79  & 17.91 & 5.73
& 64.1  & 8.09  & 19.13 & 5.91 \\

MGP
& 53.0  & 16.93 & 29.90 & 14.38
& 73.8  & 8.35  & 17.91 & 7.07
& 86.1  & 10.90 & 20.24 & 9.94 \\

\midrule

HETT
& 38.23 & 29.92 & 48.26 & 25.20
& 53.55 & 18.06 & 34.59 & 14.92
& 42.57 & 25.73 & 47.00 & 20.95 \\

RACO-Base
& 37.90 & 30.32 & 48.10 & 25.78
& 53.16 & 19.39 & 34.63 & 16.18
& 42.02 & 27.19 & 47.64 & 22.46 \\

RACO
& \textbf{36.85} & \textbf{35.55}
& \textbf{52.47} & \textbf{29.61}
& \textbf{49.96} & \textbf{27.59}
& \textbf{41.71} & \textbf{22.75}
& \textbf{40.62} & \textbf{33.71}
& \textbf{52.13} & \textbf{27.45} \\

\bottomrule
\end{tabular}
\caption{
Main navigation results on the LG-UVI episodes.
The results of Random, Human, Seq2Seq, CMA, AerialVLN, and MGP
are reported from prior work for reference.
HETT is the reproduced end-to-end reference, whereas RACO-Base
shares the backbone, scene prior, and evaluation protocol with RACO
but disables all runtime correction modules. Best automatic results are shown in \textbf{Bold}.
}
\label{tab:main_navigation_results}
\end{table*}

\section{Experiments}
\subsection{Experimental Setup}

We evaluate RACO on LG-UVI, an inspection-oriented extension of CityNav/CityRefer that retains the original instructions, trajectories, and scene maps while adding object-centric inspection annotations. We reproduce HETT as an end-to-end reference. RACO-Base uses the same two-stage architecture and Stage~1/Stage~2 policies as HETT, but its backbone is trained on LG-UVI. Full RACO further adds runtime candidate-anchor construction, coarse-goal reliability correction, and terminal inspection refinement. All three models are evaluated under the same online execution protocol.

We report the standard navigation metrics NE, SR, OSR, and SPL, together with the inspection-oriented metrics Zone-SR, OSA, ISR, and FVR. Higher values are better for SR, OSR, SPL, Zone-SR, OSA, and ISR, while lower values are better for NE and FVR. The reliability gates and terminal policies are trained on train-seen, and decision thresholds are selected on a held-out subset. The validation-seen, validation-unseen, and test-unseen splits are used only for evaluation.

RACO follows a strictly online protocol without information leakage. During inference, it uses only observable scene geometry, language-derived type cues, model candidate scores, stage status, and executed trajectory and terminal-state geometry. It never accesses ground-truth object identities or target coordinates, hard-distractor labels, final-error or rescue labels, or annotations from unseen splits, and no prediction files are modified offline. All experiments were conducted on a server with four NVIDIA RTX 4090D GPUs. Additional terminal controls and runtime statistics are reported in Supplementary Sections~S2 and S4.

\begin{table}[t]
\centering
\small
\setlength{\tabcolsep}{3.2pt}
\begin{tabular}{@{}llcccc@{}}
\toprule
Split & Model
& Zone-SR$\uparrow$
& OSA$\uparrow$
& ISR$\uparrow$
& FVR$\downarrow$ \\
\midrule
Val Seen
& HETT & 40.65 & 10.85 & \textbf{8.91} & 82.27 \\
& RACO & \textbf{45.02} & \textbf{11.17} & 7.85 & \textbf{58.14} \\
\midrule
Val Unseen
& HETT & 27.88 & 15.87 & 11.98 & 76.68 \\
& RACO & \textbf{33.96} & \textbf{16.09} & \textbf{12.72} & \textbf{57.17} \\
\midrule
Test Unseen
& HETT & 38.38 & \textbf{16.78} & \textbf{13.69} & 77.18 \\
& RACO & \textbf{44.52} & 15.87 & 12.63 & \textbf{62.09} \\
\bottomrule
\end{tabular}
\caption{Inspection-oriented diagnostics on LG-UVI under the same online protocol.}
\label{tab:inspection_diagnostics}
\end{table}

\subsection{Main Results}

Table~\ref{tab:main_navigation_results} compares RACO with existing navigation baselines on LG-UVI. Since LG-UVI retains the original CityNav episodes and navigation metrics, we include previously reported results for context. Published HETT results were obtained under a different evaluation protocol and are therefore not directly comparable. We reproduce HETT under the same online setting as RACO. RACO-Base uses the same LG-UVI-trained backbone as RACO but excludes all runtime correction modules.

RACO achieves the strongest overall navigation results among automatic methods, with the largest gains on the unseen splits. Relative to the reproduced HETT baseline, SR increases from 18.06 to 27.59 on Validation Unseen and from 25.73 to 33.71 on Test Unseen, while NE decreases from 53.55 to 49.96 and from 42.57 to 40.62, respectively. OSR and SPL follow the same trend, indicating that RACO reaches the goal region more often and completes successful episodes more efficiently. RACO also outperforms HETT on all four navigation metrics on Validation Seen.

The comparison with RACO-Base separates the effect of runtime correction from that of backbone training. Across the three splits, RACO improves SR by 5.23--8.20 percentage points, while consistently increasing OSR and SPL and reducing NE. The largest gain occurs on Validation Unseen, where SR rises from 19.39 to 27.59 and OSR from 34.63 to 41.71. This comparison shows that coarse-goal correction and terminal refinement provide benefits beyond those obtained from the LG-UVI-trained backbone alone.

The inspection-oriented diagnostics in Table~\ref{tab:inspection_diagnostics} support the same conclusion. On Validation Unseen and Test Unseen, RACO improves Zone-SR by 6.08 and 6.14 percentage points and reduces FVR by 19.51 and 15.09 points, respectively. It therefore reaches valid inspection regions more often while being less likely to verify nearby distractors. OSA and ISR show mixed results across the three splits, indicating that reliable object-level target confirmation remains a major limitation.

\begin{table*}[t]
\centering
\small
\setlength{\tabcolsep}{3.2pt}
\begin{tabular}{@{}l*{12}{c}@{}}
\toprule
\multirow{2}{*}{Variant}
& \multicolumn{6}{c}{Val Seen}
& \multicolumn{6}{c}{Val Unseen} \\
\cmidrule(lr){2-7} \cmidrule(lr){8-13}
& SR$\uparrow$ & OSR$\uparrow$ & SPL$\uparrow$ & NE$\downarrow$
& Zone-SR$\uparrow$ & FVR$\downarrow$
& SR$\uparrow$ & OSR$\uparrow$ & SPL$\uparrow$ & NE$\downarrow$
& Zone-SR$\uparrow$ & FVR$\downarrow$ \\
\midrule

RACO-Base
& 30.32 & 48.10 & 25.78 & 37.90 & 41.42 & 77.29
& 19.39 & 34.63 & 16.18 & 53.16 & 28.18 & 70.27 \\

Pre-stage only
& 31.34 & 49.43 & 26.75 & 37.44 & 42.02 & 60.24
& 21.73 & 37.12 & 18.40 & 50.97 & 30.29 & 59.49 \\

Boundary only
& 31.66 & 49.31 & 26.68 & 37.55 & 42.67 & 59.85
& 21.99 & 37.63 & 17.87 & 52.28 & 30.44 & 59.51 \\

Pre-stage + boundary
& 32.23 & 49.96 & 27.34 & 37.28 & 42.71 & 62.71
& 23.43 & 39.12 & 19.46 & 50.45 & 31.70 & 60.33 \\

Full w/o type filtering
& 32.55 & 49.03 & 27.26 & 37.27 & 42.39 & \textbf{57.13}
& 24.14 & 38.01 & 20.17 & 51.21 & 31.48 & \textbf{53.99} \\

Full RACO
& \textbf{35.55} & \textbf{52.47}
& \textbf{29.61} & \textbf{36.85}
& \textbf{45.02} & 58.14
& \textbf{27.59} & \textbf{41.71}
& \textbf{22.75} & \textbf{49.96}
& \textbf{33.96} & 57.17 \\

\bottomrule
\end{tabular}
\caption{
Ablation results on LG-UVI under the same online protocol.
``Pre-stage + boundary'' excludes terminal refinement, whereas
``Full w/o type filtering'' enables all three correction stages
without type-aware anchor filtering.
Best results for each metric are shown in \textbf{Bold}.
}
\label{tab:module_ablation}
\end{table*}

\begin{figure*}[!t]
\centering
\includegraphics[width=0.90\textwidth]{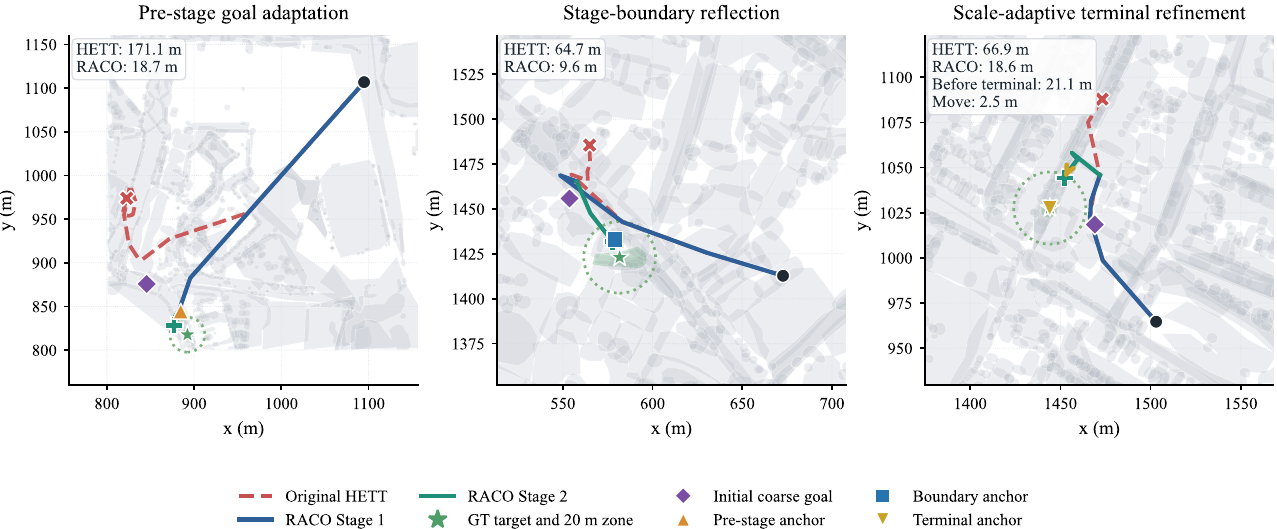}
\caption{
Qualitative correction cases on the LG-UVI development split. Each panel compares the original HETT trajectory with RACO under the same instruction. RACO corrects failures at the pre-stage, stage-boundary, and terminal refinement stages.
}
\label{fig:qualitative_cases}
\end{figure*}

\subsection{Ablation Study}

Table~\ref{tab:module_ablation} examines the contribution of each component. RACO-Base uses the LG-UVI-trained two-stage backbone with all runtime correction modules disabled and serves as the matched baseline. Both pre-stage and stage-boundary correction improve over RACO-Base, and combining them gives larger gains than either module alone. On Validation Unseen, the combined variant increases SR from 19.39 to 23.43 and OSR from 34.63 to 39.12, while also improving SPL and reducing NE. This suggests that the two modules address different errors: pre-stage correction revises unreliable initial goals, whereas stage-boundary correction repairs poor handoff states before local refinement.

Type-aware anchor filtering also contributes to performance. Removing it lowers SR, OSR, and Zone-SR on both validation splits, indicating that geometric proximity alone is insufficient for reliable anchor selection. The language-derived type cue helps reject nearby objects that do not match the instruction, improving both navigation success and inspection-region arrival. Although the variant without type filtering achieves a slightly lower FVR, it performs worse across the main navigation metrics. Full RACO therefore provides the best overall result rather than optimizing a single diagnostic measure.

Supplementary experiments show that scale-adaptive refinement outperforms the no-terminal and fixed-step variants (Table~S2), while Full RACO exceeds the strongest map-aware heuristic by 3.36 Test-Unseen SR points (Table~S3). RACO is more sensitive to object-position noise than to moderate deletion or category corruption (Table~S4), and its median Test-Unseen path-length change is zero (Table~S5).

\subsection{Qualitative Analysis}

Fig.~\ref{fig:qualitative_cases} presents representative correction cases from the LG-UVI development split. In the first case, the original HETT trajectory follows an unreliable coarse goal and remains far from the target, while RACO redirects Stage~1 toward a type-compatible runtime anchor before execution. The second case concerns an unreliable handoff at the Stage~1-to-Stage~2 boundary. Although the initial coarse navigation is partly correct, the Stage~1 endpoint is poorly aligned with the target region. RACO detects this mismatch and replans toward the selected anchor before local refinement. The third case demonstrates scale-adaptive terminal refinement. After local execution, the agent stops near the inspection region but remains outside the valid range. RACO then applies a bounded terminal move toward the runtime anchor to reach a valid final state.

These cases show how RACO corrects failures at different stages of the two-stage policy. It checks coarse-goal reliability before Stage~1, repairs unreliable handoff states before Stage~2, and corrects bounded terminal near-misses after local refinement. Ground-truth target geometry is used only for visualization and distance annotations.

\section{Conclusion}
We presented LG-UVI, an inspection-oriented extension of CityNav/CityRefer with object-centric metadata and diagnostic metrics. LG-UVI reveals a limitation of two-stage UAV-VLN policies: semantically plausible coarse goals can still be unreliable for object-level inspection, while terminal errors may leave the agent outside valid inspection regions. To address this issue, we proposed RACO, a reliability-aware coarse-to-fine framework that builds object-level candidate anchors, corrects unreliable coarse goals before Stage~1 and at the Stage~1-to-Stage~2 boundary, and applies scale-adaptive terminal refinement to near-miss cases. Experiments show that RACO improves navigation performance and inspection-region arrival while reducing false-verification risk compared with the reproduced HETT baseline. Object-level target confirmation remains difficult, motivating stronger grounding and verification for visually similar distractors.

\bibliography{aaai2027}


\newpage

\begingroup
\setcounter{secnumdepth}{2}
\setcounter{section}{0}
\setcounter{subsection}{0}
\setcounter{table}{0}
\setcounter{figure}{0}

\renewcommand{\thesection}{S\arabic{section}}
\renewcommand{\thesubsection}{S\arabic{section}.\arabic{subsection}}
\renewcommand{\thetable}{S\arabic{table}}
\renewcommand{\thefigure}{S\arabic{figure}}

\section*{Supplementary Material}
\input{supplement_body.tex}
\endgroup

\end{document}

%% file: supplement_body.tex
\section{Additional Method Details}
\label{app:method_details}

This section specifies the supervision, runtime features, estimator
configurations, acceptance rules, and bounded execution parameters used by
RACO. Ground-truth quantities are used only to construct offline training
labels and calibration statistics. They are excluded from the runtime feature
records used during navigation.

\subsection{Runtime Anchor Construction}
\label{app:runtime_anchors}

RACO uses a fixed, target-agnostic scene-level object map derived from the
CityRefer annotations. Each object entry contains a scene-level identifier,
semantic category, reference position, dimensions, and contour geometry. The
identifier is used only to index objects within a scene; the runtime map does
not indicate which object is the target of a particular episode and contains
no hard-distractor labels.

A lightweight parser normalizes explicit object names and common aliases in
the instruction into the object categories used by LG-UVI. If no category cue
can be identified, type filtering is disabled. Given the raw coarse goal
$g_0$, RACO retrieves at most $K=32$ nearby objects and constructs the runtime
anchor set $\mathcal{A}_{\mathrm{run}}$. Objects matching a language-derived
type cue are considered first, when such a cue is available. For compact
objects, the anchor reference $p(a_j)$ is the object center; for region-like
objects, it is derived from the corresponding contour geometry.

The runtime anchor set is constructed around the model-predicted goal and is
not forced to contain the ground-truth target. It is distinct from the
ground-truth-inclusive candidate pool used by the LG-UVI evaluator only after
an episode terminates. Target identifiers, target positions, ground-truth
distances, and rescue or hurt labels are never copied into the runtime feature
record.

Table~\ref{tab:runtime_feature_groups} summarizes the runtime feature groups.
Distances are represented in meters. Within-group normalized ranks and
$z$-scores are computed separately for each candidate set. The
histogram-based models use no global standardization; missing or invalid
numeric values are imputed as specified below.

\subsection{Pre-stage Goal Adapter}
\label{app:prestage_adapter}

\paragraph{Supervision.}
The pre-stage adapter is trained from frozen-backbone trajectory records and
the scene object map. Let $S_0$ indicate whether the frozen backbone terminal
state is successful under the 20~m navigation radius, and let

\begin{equation}
C_0 = \mathbf{1}\!\left[
\min_{a_j\in\mathcal{A}_{\mathrm{run}}}
d\bigl(p(a_j),p^\ast\bigr) \leq 20\ \mathrm{m}
\right],
\end{equation}

where $p^\ast$ is used only for offline label construction. Episodes are
assigned to three teacher classes: \emph{keep} when $S_0=1$,
\emph{positive correction} when $S_0=0$ and $C_0=1$, and \emph{ambiguous}
when $S_0=0$ and $C_0=0$. The intervention target is therefore

\begin{equation}
y_{\mathrm{pre}} = \mathbf{1}[S_0=0 \wedge C_0=1].
\end{equation}

For candidate ranking, a candidate is positive when it is the target object or
its reference point lies within 20~m of the target position. Exact-object,
within-radius, and negative candidate rows receive different sample weights,
allowing the binary ranker to prioritize instance-correct anchors without
exposing the target identifier at runtime.

The Train Seen records contain 21,860 labeled episodes: 10,002 positive
corrections, 9,255 keep cases, and 2,603 ambiguous cases. The corresponding
candidate table contains 319,780 rows, including 8,529 exact-object rows,
41,006 additional within-20-m rows, and 270,245 negative rows. The ranker is
fitted only on positive-correction episodes.

\begin{table}[!t]
\centering
\scriptsize
\setlength{\tabcolsep}{3pt}
\begin{tabular}{@{}p{0.22\columnwidth}c p{0.60\columnwidth}@{}}
\toprule
Module & Dim. & Main runtime feature groups \\
\midrule
Pre-stage gate
& 24
& Candidate distance statistics, type counts, type cue \\
Pre-stage ranker
& 47
& Goal/agent geometry, type match, local density, group statistics \\
Boundary gate
& 118
& Stage-1 trajectory, endpoint geometry, candidate-score distribution \\
Terminal scorer
& 46
& Terminal geometry, proposed action, recent progress, gate diagnostics \\
\bottomrule
\end{tabular}
\caption{Runtime feature groups used by the learned decision modules. The
dimension excludes ground-truth-only quantities used for offline labeling.}
\label{tab:runtime_feature_groups}
\end{table}

\paragraph{Models and objective.}
The gate and pointwise candidate ranker are binary
\texttt{HistGradientBoostingClassifier} models optimized with weighted
log-loss. Both use 160 boosting iterations, learning rate $0.06$,
$\ell_2$ regularization $0.02$, minimum leaf size 25, and random state 17.
For the gate, the teacher-class weights are 2.5 for keep, 1.2 for ambiguous,
and 1.0 for positive correction. For the ranker, exact-object,
within-20-m, and negative rows receive weights 4.0, 2.0, and 0.5,
respectively. Invalid numeric values are replaced by zero before fitting.

Let $\rho_{\mathrm{pre}}$ denote the gate probability, $u_j$ the candidate
probability, $a^\star=\arg\max_j u_j$, and
$\Delta_u=u_{(1)}-u_{(2)}$ the top-two score margin. The final pre-stage
acceptance rule is

\begin{equation}
\Gamma_{\mathrm{pre}} =
\mathbf{1}\!\left[
\begin{aligned}
&\rho_{\mathrm{pre}} \geq 0.55,\quad
u(a^\star) \geq 0.05,\quad \Delta_u \geq 0,\\[-1mm]
&\|p(a^\star)-g_0\|_2 \leq 80\ \mathrm{m}
\end{aligned}
\right].
\end{equation}

If $\Gamma_{\mathrm{pre}}=1$, Stage~1 uses
$g'_0=p(a^\star)$; otherwise it retains $g_0$. The calibration sweep uses gate
and rank thresholds from $\{0.05,0.10,\ldots,0.95\}$ and margin thresholds
from $\{0,0.01,0.03,0.05,0.08,0.12,0.18,0.25\}$. Among configurations
satisfying the prescribed no-harm constraint on originally successful
calibration episodes, selection prioritizes proxy success, net rescue, and
selected-anchor precision. The selected calibration record contains 124
accepted interventions, 46 proxy rescues, and 7 proxy hurts.

\subsection{Stage-boundary Reliability and Coarse Replanning}
\label{app:boundary_replanning}

\paragraph{Reliability supervision.}
At the Stage~1-to-Stage~2 boundary, the learned gate predicts whether the
current handoff is likely to lead to a large terminal error. Its offline
binary target is

\begin{equation}
y_{\mathrm{B}}=\mathbf{1}[\mathrm{NE}_{\mathrm{final}}>25\ \mathrm{m}],
\end{equation}

where final NE is used only for label construction. The runtime feature vector
contains Stage~1 trajectory length, displacement, straightness and turning
statistics; endpoint geometry relative to the predicted goal and candidate
anchors; candidate counts, type composition, entropy and spatial spread; and
the top-score, margin, entropy and rank statistics of the runtime candidate
scores. Ground-truth endpoint error, candidate NE, final NE, and rescue or hurt
indicators are excluded. The training table contains 21,877 episodes, with
10,371 positive and 11,506 negative labels.

The boundary gate applies median imputation followed by a binary
\texttt{HGB classifier} with 160 boosting iterations,
learning rate $0.05$, maximum 31 leaf nodes, $\ell_2$ regularization $0.02$,
and random state 17. Threshold selection sweeps all distinct calibration
probabilities and retains settings satisfying precision at least 0.75,
no-harm at least 0.975, and intervention coverage between 0.05 and 0.12. The
frozen threshold is 0.7934106296. At this operating point, the calibration
record contains 161 predicted unreliable states, with precision 0.8447,
no-harm 0.9752, and coverage 0.0652.

\paragraph{Candidate selection and bounded replanning.}
When the boundary probability exceeds the frozen threshold, the executor uses
the \texttt{top\_score} picker and the
\texttt{lg\_uvi\_choice\_scores} field to select the highest-scoring
admissible anchor from at most 32 runtime candidates. An anchor is admissible
only when its current distance is at most 80~m. The selected reference
$p(a_{\mathrm{B}}^\star)$ replaces the Stage~2 coarse goal, after which RACO
performs a bounded coarse replan. Replanning
terminates when the UAV is within 8~m of the selected anchor or after eight
additional coarse steps. Each extra coarse step uses
\texttt{move\_iteration} equal to 10 in the navigation executor, while the
episode-level maximum action length remains 20. If the gate does not activate
or no admissible candidate is available, Stage~2 starts from the original
handoff state without boundary correction.

\subsection{Scale-adaptive Terminal Inspection Refinement}
\label{app:terminal_policy_details}

\paragraph{Action basis and supervision.}
Let $a_T$ be the runtime inspection anchor supplied by the pre-stage anchor
module and
$d_T=\|x_T-p(a_T)\|_2$. Terminal refinement is considered only for
$20\leq d_T\leq35$~m. The geometry-derived base move is

\begin{equation}
m_{\mathrm{geo}}(d_T)=
\begin{cases}
5,  & 20\leq d_T<25,\\
10, & 25\leq d_T<30,\\
15, & 30\leq d_T\leq35,\\
0,  & \text{otherwise}.
\end{cases}
\end{equation}

For $\alpha\in\mathcal{S}=\{0,0.5,0.75,1.0,1.25\}$, the corresponding
counterfactual endpoint moves from $x_T$ toward $p(a_T)$ by
$\alpha m_{\mathrm{geo}}(d_T)$, capped by the anchor distance. The positive
training target is a \emph{rescue}: the original endpoint is unsuccessful but
the counterfactual endpoint falls within the 20~m success radius. Hurt,
no-rescue, and unneeded-safe actions are negative targets and receive separate
weights. This construction is an offline counterfactual proxy; the navigation
backbone is not rerun separately for every training action.

The training set contains 36,630 action rows from 7,326 eligible episodes:
3,993 rescue, 1,694 hurt, 18,797 no-rescue, and 12,146 unneeded-safe rows.
Their sample weights are 6.0, 9.0, 0.7, and 0.4, respectively.

\paragraph{Action scorer and runtime rule.}
The terminal module is a binary action scorer rather than a categorical
softmax policy. It uses a \texttt{HGB classifier} with 180
boosting iterations, learning rate $0.04$, $\ell_2$ regularization $0.05$,
minimum leaf size 25, random state 23, and weighted log-loss. Its 46 runtime
features summarize terminal-to-anchor geometry, recent trajectory progress,
last-step length, geometry bucket, proposed movement and scale, candidate
ambiguity, and pre-stage and boundary diagnostics. Missing numeric values are
replaced by zero.

At runtime, the scorer evaluates every $\alpha\in\mathcal{S}$ independently
and selects the largest score. A zero-scale action represents abstention. The
frozen score threshold is 0, the eligible anchor-distance interval is
$[20,35]$~m, and the selected calibration configuration satisfies an
application-rate bound of 32.1\% and a minimum no-harm constraint of 0.937.
The accepted movement is capped by both the selected scale action and the
current anchor distance and is applied before the final online metrics are
computed.

\subsection{Candidate Selector}
\label{app:candidate_selector}

The external candidate selector is trained on samples for which the target
object occurs in a candidate set of at most 32 same-type objects. Target
identifiers are used to construct training labels but are not supplied at
runtime. Each candidate token combines a 512-dimensional CLIP image feature,
a 512-dimensional CLIP text feature, their elementwise product and absolute
difference, image--text similarity and similarity rank, 15 geometry features,
24 language/map features, and a 12-dimensional object-type embedding. The
projected tokens are processed by a two-layer Transformer encoder with hidden
dimension 256 and four attention heads, followed by text cross-attention and
an MLP score head. The selector contains 2,683,303 trainable parameters.

For target candidate $j^\ast$, the selector objective is

\begin{equation}
\mathcal{L}_{\mathrm{sel}}
=\mathcal{L}_{\mathrm{CE}}
+0.35\,\mathcal{L}_{\mathrm{hard}}(0.25)
+0.25\,\mathcal{L}_{\mathrm{top}k}(k=5,0.12),
\end{equation}

where the auxiliary terms enforce margins against the hardest negative and
the top-$k$ negative set. When type-balanced training is enabled, the
cross-entropy weights for Building, Car, Ground, and Parking are 1.5, 2.2,
1.0, and 1.0. The reported configuration uses batch size 16, eight epochs,
AdamW with learning rate $2\times10^{-4}$ and weight decay $10^{-4}$,
dropout 0.35, a hybrid full-instruction plus target-phrase text input, top-32
candidates, $224\times224$ image patches, and altitude 80~m. A single selector
checkpoint is frozen before unseen-split evaluation.

\subsection{Training, Calibration, and Leakage Protocol}
\label{app:training_protocol}

All learned gates and scorers are fitted using Train Seen records. Model
selection and threshold calibration use the designated calibration split in
the experimental protocol; all models, thresholds, candidate-selection rules,
and execution bounds are then frozen before unseen-split evaluation. The
calibration search changes only decision thresholds and does not refit the
navigation backbone.

Ground-truth object identifiers, target coordinates, hard-distractor labels,
ground-truth endpoint errors, final NE, candidate NE, rescue or hurt labels,
and split identifiers may be used to construct offline labels or evaluation
statistics, but none is included in runtime features. Runtime decisions use
only the instruction, initial and current poses, executed trajectory history,
model-predicted goals and candidate scores, language-derived type cues, and
the target-agnostic scene object map. The ground-truth-inclusive LG-UVI
candidate pool is invoked only after navigation terminates to compute the
diagnostic metrics in Section~\ref{app:metric_definitions}.

\subsection{Protocol for Additional Analyses}
\label{app:additional_analysis_protocol}

All supplementary controls use the same LG-UVI data splits and online metric
computation as the main paper. RACO-Base denotes the same LG-UVI-trained
two-stage backbone with all external runtime corrections disabled. For the
object-map robustness analysis, only the runtime map is perturbed; the
instruction, initial state, backbone outputs, frozen RACO model, and metric
computation remain unchanged. Object deletion removes the stated fraction of
map entries, position perturbation modifies object reference positions at the
stated magnitude, and category perturbation corrupts the stated fraction of
semantic labels.

\section{Additional Terminal Refinement Ablation}
\label{app:terminal_ablation}

Table~\ref{tab:terminal_ablation} isolates the terminal module while
keeping the pre-stage and stage-boundary correction modules unchanged.

\begin{table}[!ht]
\centering
\scriptsize
\setlength{\tabcolsep}{2.0pt}
\begin{tabular}{@{}llccccc@{}}
\toprule
Split & Variant
& SR$\uparrow$ & SPL$\uparrow$ & NE$\downarrow$
& Zone-SR$\uparrow$ & ISR$\uparrow$ \\
\midrule
Val Seen & No terminal
& 32.23 & 27.34 & 37.28 & 42.71 & 7.53 \\
& Action len. 21
& 32.02 & 26.98 & 37.39 & 42.87 & 7.57 \\
& Fixed 5 m
& 32.91 & 27.89 & 37.23 & 42.91 & 7.57 \\
& Scale-adaptive
& \textbf{35.55} & \textbf{29.61} & \textbf{36.85}
& \textbf{45.02} & \textbf{7.85} \\
\midrule
Val Unseen & No terminal
& 23.43 & 19.46 & 50.45 & 31.70 & 12.38 \\
& Fixed 5 m
& 24.43 & 20.33 & 50.36 & 32.18 & 12.46 \\
& Scale-adaptive
& \textbf{27.59} & \textbf{22.75} & \textbf{49.96}
& \textbf{33.96} & \textbf{12.72} \\
\bottomrule
\end{tabular}
\caption{
Terminal-refinement ablation on LG-UVI. All variants use the same
pre-stage and stage-boundary correction modules. ``Action len. 21''
increases the execution budget without applying terminal correction,
whereas ``Fixed 5 m'' applies a constant terminal displacement. Best
results within each split are shown in bold.
}
\label{tab:terminal_ablation}
\end{table}

On Validation Seen, scale-adaptive refinement improves SR from 32.23 to
35.55, SPL from 27.34 to 29.61, and Zone-SR from 42.71 to 45.02, while
reducing NE from 37.28 to 36.85. On Validation Unseen, it improves SR
from 23.43 to 27.59 and SPL from 19.46 to 22.75, while reducing NE from
50.45 to 49.96. The fixed 5 m displacement provides smaller gains, and
increasing the action budget on Validation Seen does not improve SR over
the no-terminal variant. These controls indicate that the observed gain
is not explained solely by longer execution or by the tested fixed
displacement. Among the settings evaluated in
Table~\ref{tab:terminal_ablation}, the scale-adaptive policy provides the
strongest overall results.

\section{Map-Aware Controls and Object-Map Robustness}
\label{app:map_controls}

\subsection{Map-Aware Heuristic Baselines}

To examine whether access to the object map alone explains the
improvement, we compare Full RACO with non-learning heuristics that use
the same navigation backbone, annotation-derived scene prior, and online
evaluation protocol.

\begin{table}[!ht]
\centering
\small
\begin{tabular}{@{}lc@{}}
\toprule
Method & Test Unseen SR$\uparrow$ \\
\midrule
Nearest-type heuristic & 27.70 \\
Nearest-type + geometry terminal & 30.35 \\
Full RACO & \textbf{33.71} \\
\bottomrule
\end{tabular}
\caption{
Test Unseen SR for Full RACO and non-learning controls using the same
annotation-derived object map.
}
\label{tab:map_heuristics}
\end{table}

Full RACO exceeds the strongest tested map-aware heuristic by 3.36
percentage points. Thus, object-map access together with the tested
nearest-type and geometry-terminal rules does not account for the full
improvement of RACO.

The heuristic controls retain the same anchor proposal space and
geometric execution bounds but remove RACO's learned reliability and
utility models. Their lower performance therefore shows that structured
map access and geometric correction alone do not account for the gain.

\subsection{Synthetic Object-Map Perturbations}

We further evaluate sensitivity to synthetic perturbations of the
annotation-derived object map on Validation Unseen. To keep this
controlled analysis separate from the absolute main-table comparison,
Table~\ref{tab:map_robustness} reports the paired change in SR relative
to the clean-map run from the same perturbation evaluation.

\begin{table}[!ht]
\centering
\small
\setlength{\tabcolsep}{6pt}
\begin{tabular}{@{}lcc@{}}
\toprule
Map condition & SR$\uparrow$ & $\Delta$SR \\
\midrule
Clean map & 27.59 & 0.00 \\
Delete 10\% objects & 27.29 & -0.30 \\
Delete 20\% objects & 26.88 & -0.71 \\
Delete 30\% objects & 26.59 & -1.00 \\
Position noise (5 m) & 24.10 & -3.49 \\
Position noise (10 m) & 23.58 & -4.01 \\
Category noise (10\%) & 26.40 & -1.19 \\
Category noise (20\%) & 26.40 & -1.19 \\
\bottomrule
\end{tabular}
\caption{
Validation Unseen SR under synthetic object-map perturbations. The
reported change is measured relative to the clean-map condition.
}
\label{tab:map_robustness}
\end{table}

Object deletion and moderate category corruption produce relatively
small SR changes under the tested perturbations. In contrast, perturbing
object reference positions by 5 or 10 m causes a larger performance
drop, indicating that RACO is more sensitive to localization error than
to moderate object omission or category noise. These experiments use
synthetic perturbations of an annotation-derived scene prior and should
not be interpreted as an evaluation with a learned or online-predicted
object map.

\section{Runtime Behavior and Model Cost}
\label{app:runtime_cost}

Table~\ref{tab:trigger_rates} reports how often each correction stage is
activated on Test Unseen. A module is counted as triggered when it is
activated at least once in an episode; the reported rate is the percentage
of Test Unseen episodes satisfying this condition.

\begin{table}[!ht]
\centering
\small
\begin{tabular}{@{}lc@{}}
\toprule
Runtime module & Test trigger rate \\
\midrule
Pre-stage correction & 8.9\% \\
Boundary correction & 10.5\% \\
Terminal refinement & 29.9\% \\
\bottomrule
\end{tabular}
\caption{
Activation rates of the three RACO correction stages on Test Unseen.
}
\label{tab:trigger_rates}
\end{table}

Relative to RACO-Base, Full RACO increases the average Test Unseen path
length by 3.98 m, while the median path-length difference is zero. Among
5,281 episodes, 2,883 have unchanged path length. These results show
that the runtime modules intervene selectively rather than modifying
every trajectory.

The candidate-anchor selector contains 2,683,303 parameters. The selector
and the three lightweight decision-module files together require
approximately 13.2 MB of storage.

\section{Formal Definitions of LG-UVI Diagnostics}
\label{app:metric_definitions}

Let $N$ denote the number of evaluation episodes, $x_T^i$ the terminal
position in episode $i$, $o_i^\ast$ the target object, $\hat{o}_i$ the
object associated with the terminal prediction by the evaluation
protocol, and $Z(o_i^\ast)$ the type-aware inspection region. The four
inspection-oriented diagnostics are defined as
\[
\begin{aligned}
\mathrm{Zone\mbox{-}SR} &=
\frac{1}{N}\sum_{i=1}^{N}\mathbf{1}[x_T^i \in Z(o_i^\ast)],\\
\mathrm{OSA} &=
\frac{1}{N}\sum_{i=1}^{N}\mathbf{1}[\hat{o}_i = o_i^\ast],\\
\mathrm{ISR} &=
\frac{1}{N}\sum_{i=1}^{N}
\mathbf{1}[x_T^i \in Z(o_i^\ast)]
\mathbf{1}[\hat{o}_i = o_i^\ast],\\
\mathrm{FVR} &=
\frac{1}{N}\sum_{i=1}^{N}
\mathbf{1}[\hat{o}_i \in \mathcal{D}_h(o_i^\ast)] .
\end{aligned}
\]
Here $\mathcal{D}_h(o_i^\ast)$ denotes the hard-distractor set associated
with the target object. Zone-SR measures inspection-region arrival, OSA
measures object association accuracy, ISR requires both correct region
arrival and correct object association, and FVR measures association
with a hard distractor. All four metrics are averaged over the same $N$
evaluation episodes. When a method produces an explicit object candidate,
the evaluator uses its predicted identifier as $\hat{o}_i$; otherwise, it
applies the deterministic same-type association rule used by the LG-UVI
evaluation protocol. If no object can be associated, $\hat{o}_i$ is set to
the empty symbol and contributes zero to the object-association indicators.
The ground-truth-inclusive evaluation pool is used only to compute these
post-hoc diagnostics and is never exposed to RACO during trajectory
execution.